\documentclass[final]{cvpr}

\usepackage{times}
\usepackage{epsfig}
\usepackage{graphicx}
\usepackage{amsmath}
\usepackage{amssymb}

\usepackage{booktabs}
\usepackage{subfigure}
\usepackage{diagbox}
\usepackage{makecell}
\usepackage{multirow}

\newtheorem{theorem}{Theorem}

\usepackage[pagebackref=true,breaklinks=true,colorlinks,bookmarks=false]{hyperref}

\begin{document}

\title{Improving Robustness for Pose Estimation via Stable Heatmap Regression}

\author{Yumeng Zhang\\
\small School of Software, BNRist,\\
\small Tsinghua University\\
\small Beijing, China\\
\and
Li Chen \thanks{Corresponding author: Li Chen}\\
\small School of Software, BNRist,\\
\small Tsinghua University\\
\small Beijing, China\\
\and
Yufeng Liu\\
\small SEU-ALLEN Joint Center, \\
\small Southeast University\\
\small Nanjing, China\\
\and
Xiaoyan Guo\\
\small Y-tech, \\
\small Kuaishou Technology\\
\small Beijing, China\\
\and
Wen Zheng\\
\small Y-tech, \\
\small Kuaishou Technology\\
\small Beijing, China\\
\and
Junhai Yong\\
\small School of Software, BNRist,\\
\small Tsinghua University\\
\small Beijing, China\\
}



\maketitle

\begin{abstract}
   Deep learning methods have achieved excellent performance in pose estimation, but the lack of robustness causes the keypoints to change drastically between similar images. In view of this problem, a stable heatmap regression method is proposed to alleviate network vulnerability to small perturbations. We utilize the correlation between different rows and columns in a heatmap to alleviate the multi-peaks problem, and design a highly differentiated heatmap regression to make a keypoint discriminative from surrounding points. A maximum stability training loss is used to simplify the optimization difficulty when minimizing the prediction gap of two similar images. The proposed method achieves a significant advance in robustness over state-of-the-art approaches on two benchmark datasets and maintains high performance.
\end{abstract}

\section{Introduction}

\begin{figure}[tb]
   \centering
   \includegraphics[width=0.95\linewidth]{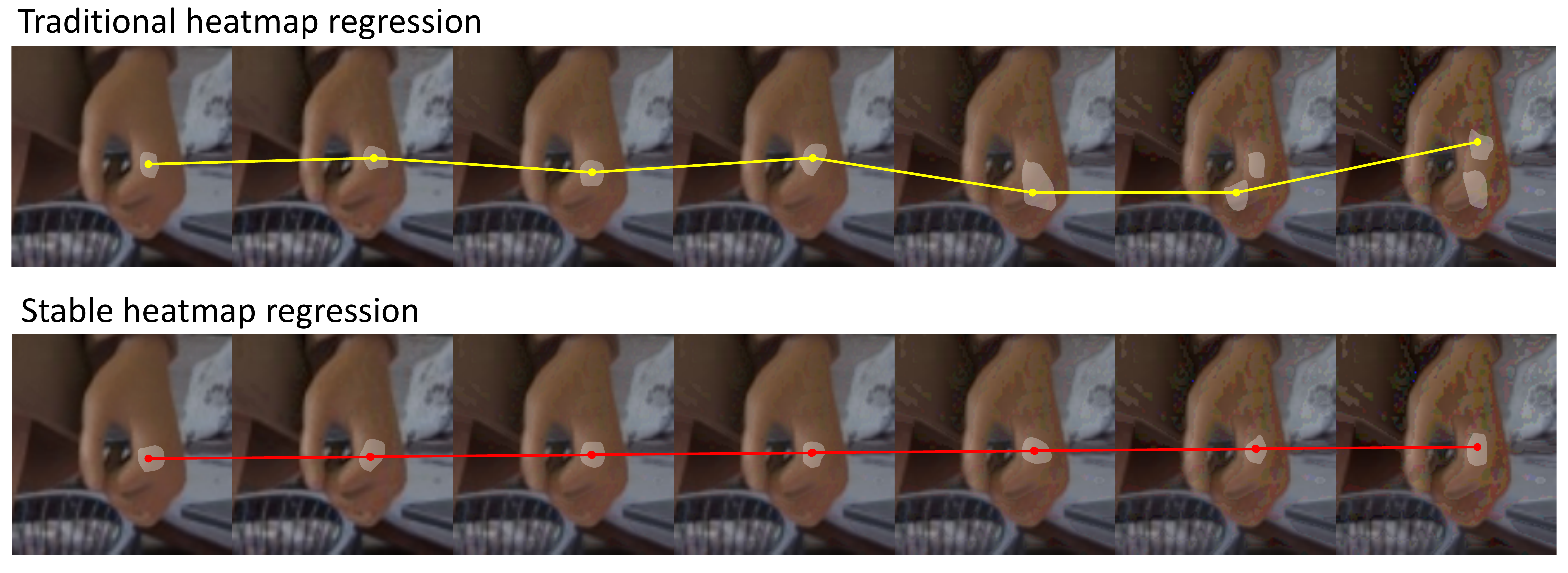}
   \caption{The predicted keypoints of traditional heatmap regression and those of the proposed stable heatmap regression on seven similar images. The translucent white parts represent areas of high responses on the heatmap and the points are the predicted keypoints. These images are almost same for humans, whereas the traditional method produces unstable predictions (yellow points). The proposed stable heatmap regression is going to alleviate this problem.} 
   \label{fig:breifintroduction}
 \end{figure}

With the continuous development of deep learning, the performance of pose estimation has been improved in unprecedented ways, thus increasing its wide application in human-computer interaction, virtual reality and many other fields. However, existing pose estimation models can be extremely vulnerable to small perturbations despite their excellent performance. These small perturbations do not change the semantic information of an image, but they can greatly alter the output of the networks. For example, the predictions of keypoints change considerably in similar continuous frames, as illustrated in Fig \ref{fig:breifintroduction}. This problem greatly affects the user experience. Therefore, improving the robustness of a pose estimation network is an urgent research task.

Improving the robustness of a network is challenging due to the black-box nature of deep learning methods. At present, few papers have analyzed the stability of pose estimation task. Most of the existing studies for improving robustness focus on a relatively simple task-image classification, such as famous adversarial examples \cite{goodfellow2014explaining}, and a large number of robustness-enhancing methods \cite{carmon2019unlabeled,madry2017towards,tramer2019adversarial,yin2019fourier,zhang2019theoretically} have emerged in classification to improve the robustness of networks against small perturbations. 

However, most of these robustness-enhancing methods are unsuitable for pose estimation. For example, Mixup \cite{zhang2017mixup} and Cutmix \cite{yun2019cutmix} improved the robustness by performing data augmentations, but these methods may cause occlusions of keypoints, resulting in meaningless training samples. Some other robustness-enhancing strategies, although theoretically applicable to different tasks, have limited or even negative effects on robustness in pose estimation task because of their characteristic. For example, limiting the Lipschitz of networks \cite{usama2018towards} or using the largest $k$ values activation function \cite{xiao2019resisting} significantly improve robustness against adversarial attacks and scale to pose estimation theoretically. However, these approaches limit the expressive power of networks while pose estimation models rely on rich global and local features to determine the position of keypoints. Thus, when these methods are applied to pose estimation task, the network fails to converge or the performance becomes unsatisfactory. To the best of our knowledge, few methods have been proposed to analyze or improve the robustness of the pose estimation models. The current study is going to fill this gap. 

Initially, we analyze the instability factors of pose estimation task, and then design corresponding modules to improve the robustness. The robustness analysis is based on the heatmap regression, a commonly used method in pose estimation. The heatmap is generally the same size as the input image and each value represents the probability of the corresponding point to belong to a keypoint in the input image. The point with the highest probability is often identified as the predicted keypoint. Many methods have achieved high performance with this strategy, but several robustness problems were ignored. We believe that the networks are not robust with the current heatmap regression methods because of three reasons.

 \textbf{The first is multi-peaks predictions.} When predicting keypoints of some difficult samples, similar high responses may appear in different regions of the heatmap. Thus, a slight change in the input may cause the predicted keypoint to shift from one peak to another. This problem has been illustrated in the two rightmost images in the first row of Fig. \ref{fig:breifintroduction}.

 \textbf{The second is the use of Gaussian heatmap as ground truth.} Current ground-truth heatmap is obtained by a Gaussian distribution. The maximum and second-largest value of an output heatmap are too close in value due to the use of this Gaussian heatmap. Thus, small perturbations in the output may cause the change of predicted keypoints. 

\textbf{The third is the optimization difficulty when combining heatmap regression with existing stability training (ST) methods.} Many ST methods \cite{laermann2019achieving,zhang2019theoretically,zheng2016improving} improve robustness by minimizing the distance between the output of a clean image and its perturbed image. However, the output probabilities of a heatmap are different from those in the classification. The output heatmap represents the probability that each pixel in the image belongs to a keypoint, and each pixel of a clean image may be changed in its perturbed image. Forcing the probabilities of all the pixels between these image pairs to be the same increases difficulty in network optimization. 

In view of these problems, we propose a stable heatmap regression, an alternative to traditional heatmap regression, with three novel robustness-enhancing designs.

\begin{itemize}

\item [1)] \textbf{A Row-Column Correlation (RCC) layer} to alleviate the multi-peaks problem. This layer utilizes the correlation between different rows and columns in the heatmap to detect the multi-peaks problem. Then it helps the networks output single-peak heatmap by increasing the gap between different peaks and raising the loss of multi-peaks heatmap.
 
\item [2)] \textbf{A Highly Differentiated Heatmap Regression (HDHR) method} to make a keypoint prediction discriminative from surrounding points. We replace the Gaussian heatmap with a multi-label map and a well-designed highly differentiated heatmap. Then we design a weighted cross-entropy loss based on them to enlarge the difference between maximum and the second largest value in the heatmap.

\item [3)] \textbf{A Maximum Stability Training (MST) loss} to simplify the optimization difficulty when minimizing the gap between the outputs of two similar images. This loss focuses on changes of the position with the largest value in the heatmap. And we provide a theoretical certification to demonstrate the effectiveness of MST.
\end{itemize}

The proposed method is evaluated on two datasets with six different model architectures. The experimental results demonstrate that our method enhances the robustness of pose estimation models and achieves a significant advance over existing methods for improving robustness. 

\begin{figure*}[tb]
  \centering
  \includegraphics[width=0.97\linewidth]{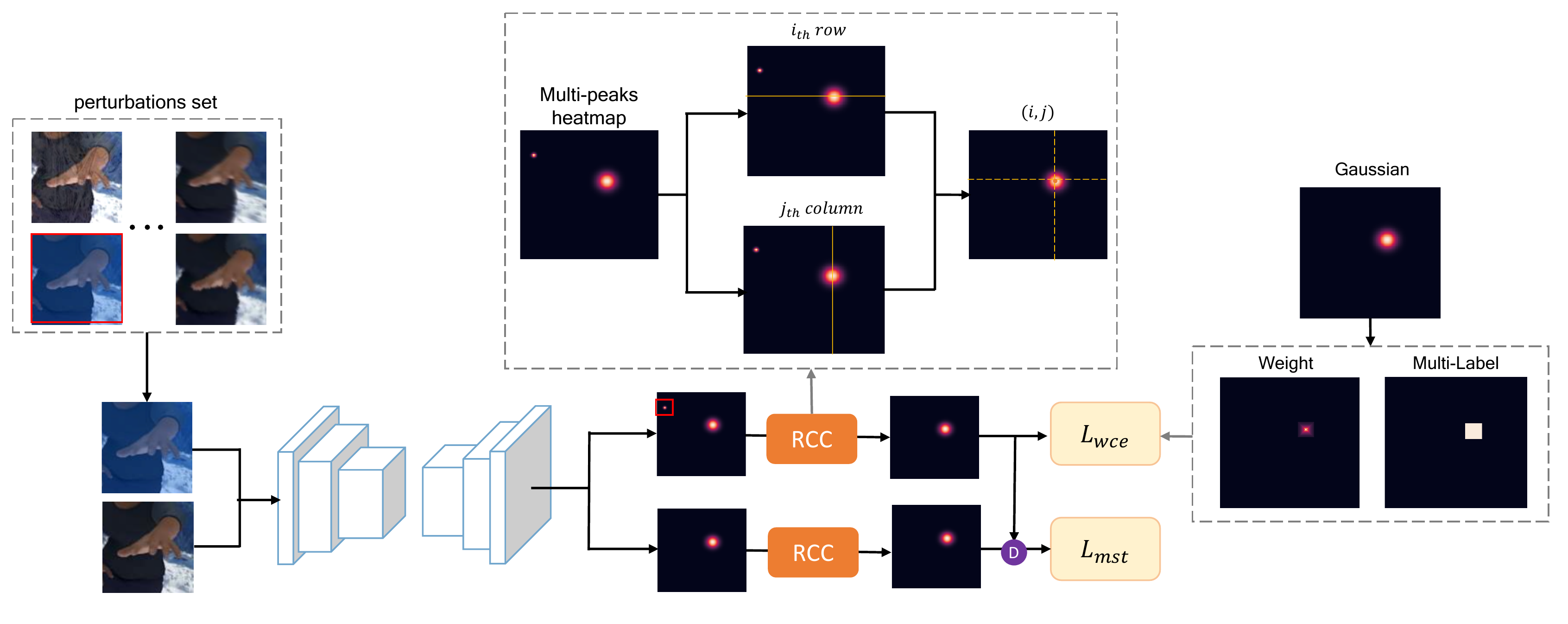}
  \caption{Schematic of our stable heatmap regression algorithm. The proposed method takes a clean image and its perturbed image as input. The perturbed image is selected randomly from a pre-defined perturbation set. A Row-Column Correlation (RCC) layer is designed to alleviate the multi-peaks problem. Traditional L2 or L1 loss on Gaussian heatmap are replaced with weighted cross-entropy loss $\mathcal L_{wce}$ on a multi-label map, where the highly differentiated heatmap is used as the weight of the multi-label map. The difference (D) between the outputs of two input images is used to calculate the Maximum Stability Training loss $\mathcal L_{mst}$.} 
  \label{fig:MPT}
\end{figure*} 

\section{Related works}

Very few works in the literature have explored ways to improve the robustness for pose estimation. The task, however, shares several properties with adversarial examples \cite{goodfellow2014explaining,goldblum2020adversarially} and natural corruptions \cite{hendrycks2019benchmarking,rusak2020simple}. Some approaches that have been proposed for these fields can be adapted for pose estimation. Hence, in the following section, we discuss the related works for improving the robustness of networks in general.

{\bf Stability training} Stability training (ST) \cite{kannan2018adversarial,zheng2016improving} is an effective strategy to improve robustness. Zheng et al. \cite{zheng2016improving} first proposed a regularization loss to make the features or outputs of a clean image and its perturbed image consistent. Li et al. \cite{li2019certified} further provided a theoretical evidence for Zheng's loss by analyzing the upper bound on the tolerable size of perturbations with Renyi divergence \cite{van2014renyi} and Gaussian noise. Subsequently, a variant of Zheng's \cite{zheng2016improving} loss is proposed by Zhang et al. \cite{zhang2019theoretically}. They replaced the L2 metric with the KL-divergence metric on the basis of their theoretical analysis. More recent studies by Hendrycks et al. \cite{hendrycks2019augmix} and Lopes et al. \cite{lopes2019improving} focused on the construction of the perturbation set. Hendrycks et al. \cite{hendrycks2019augmix} obtained more general perturbations by combining different augmentation methods and then applied Zheng's loss \cite{zheng2016improving} on these perturbations to improve robustness. Lopes et al. \cite{lopes2019improving} proposed a patch gaussian augmentation method to improve robustness and retain high performance on clean images. ST methods can be easily adapted to different tasks. However, task-specific modifications should be considered to further improve the robustness on a certain task.

{\bf Network robustness analysis} Many methods have been proposed to identify the component that causes instability through the analysis of the network property or architecture. For example, Tsuzuku et al. \cite{NIPS2018_7889} and Usama et al. \cite{usama2018towards} have found that networks with smaller Lipschitz constant are more robust. Then, they proposed corresponding strategies to limit this constant. However, the Lipschitz constant of networks is related to the expressive power of them. Limiting Lipschitz degrades the performance on complex tasks or hard datasets. Meanwhile, Azulay et al. \cite{azulay2018deep} discovered that the downsampling operation caused the lack of translation invariant in convolutional neural networks (CNN). However, downsampling cannot be easily removed from CNN models, because the shape of the output is not always consistent with that of the input. Hendrycks et al. \cite{hendrycks2019benchmarking} pointed out that the tricks, such as multiscale architectures and larger feature aggregating models, could improve the robustness, whereas these tricks were obtained by experiments without considering the underlying causes of the instabilities.     

To conclude, the improvement of robustness in pose estimation is still an underexplored problem. Our work is going to fill this gap.

\section{Method}

The proposed stable heatmap regression is shown in Fig. \ref{fig:MPT}. It consists of three components, row-column correlation layer, highly differentiated heatmap regression and maximum stability training. The three modules are used to alleviate the multi-peaks problem, enlarge the difference between maximum and the second-largest value of a heatmap, and limit the prediction gap between a clean image and its perturbed image respectively. Next, we will introduce the three modules in detail.

\subsection{Row-column correlation layer}

\begin{figure}[tbp]
  \centering
  \subfigure[Variance Case]{
     \includegraphics[width=0.29\linewidth]{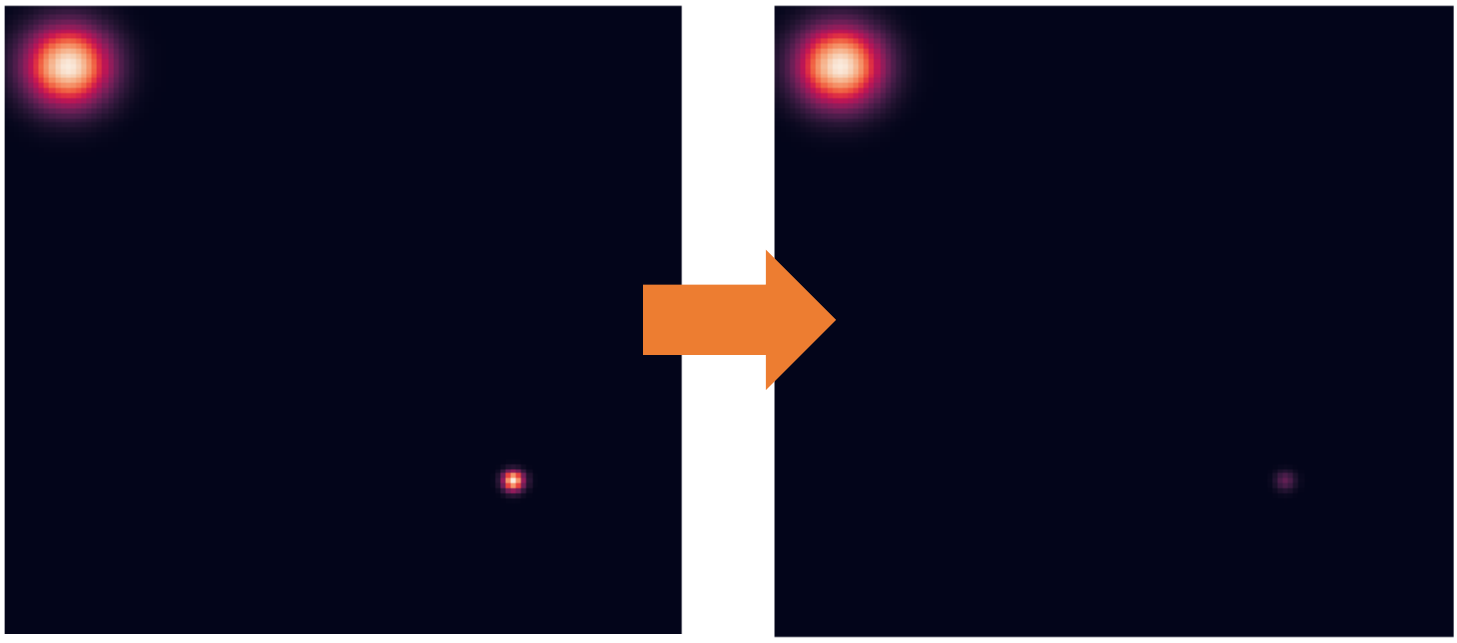}
  }
  \subfigure[Center Case]{
     \includegraphics[width=0.29\linewidth]{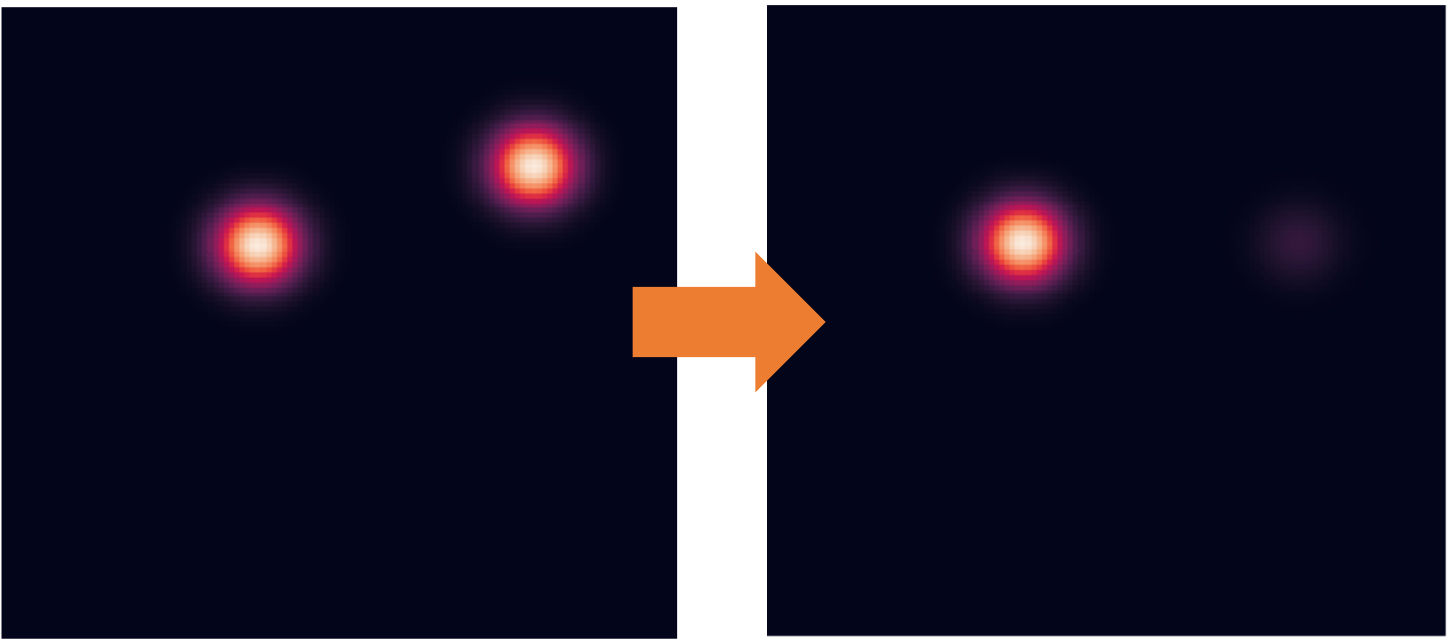}
  }
  \subfigure[Equal Case]{
    \includegraphics[width=0.29\linewidth]{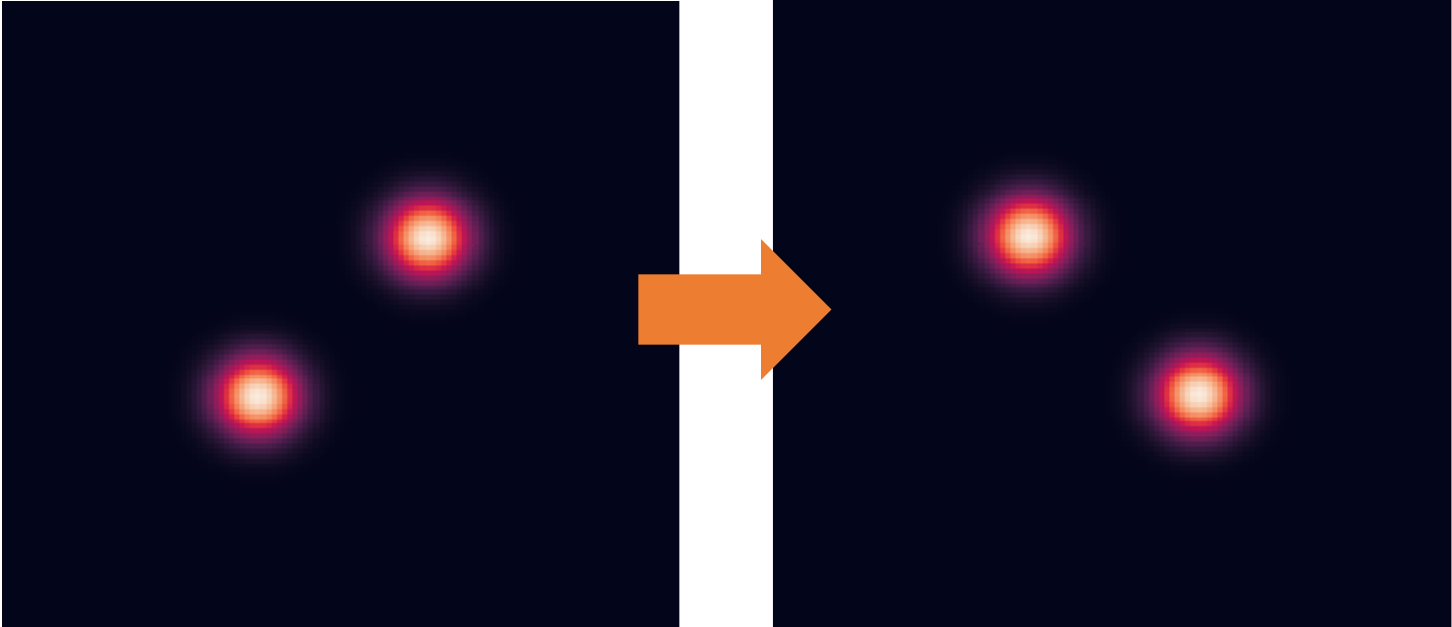}
 }
  \caption{The role of the proposed RCC layer in reducing multi-peaks problem. Cases in (a) and (b) illustrate that RCC layer tends to attenuate peaks with small variance and far from the center of the heatmap. Case (c) shows a rare situation that $C_1$ and $C_2$ (defined by Eq. (\ref{eq:jianhua})) are equal. In this case, the RCC layer generates two peaks on the diagonal. The two peaks are usually wrong and result in a big loss. Thus, the network could realize that the single-peak output is better as the loss decreases. 
  }

\label{fig:spcases}
\end{figure}

Row-Column Correlation (RCC) layer is designed to alleviate the multi-peaks phenomenon in the predicted heatmaps. The layer can detect the multi-peaks problem and alleviate it by increasing the gap between different peaks and raising the loss of multi-peaks heatmap. In this section, we first introduce the operation of this layer and then provide a theoretical analysis to explain its behavior. The operation of $RCC$ is shown in (\ref{eq:rcoper}).

\begin{equation}
  RCC(l,m) = {\sum_{i=1}^W HT(l,i)HT(i,m)}
  \label{eq:rcoper}
\end{equation}
where $RCC(l,m)$ is the output of Row-Column Correlation layer at $l_{th}$ row and $m_{th}$ column, $HT$ is the input heatmap of RCC and $W$ is the width of $HT$. Through this definition, the value at $(l,m)$ of $RCC$ represents the correlation between $l_{th}$ row and $m_{th}$ column of the input heatmap. RCC layer requires that the input heatmap has the same height and width, and this can be easily achieved by cropping or padding. 

A theoretical analysis of $RCC$ layer is provided based on a N-peak Gaussian heatmap to demonstrate the elegant properties of $RCC$. The N-peak heatmap is defined as the sum of N independent Gaussian sub-heatmaps, where the center and variance of $n_{th}$ Gaussian heatmap are $(kr_n,kc_n)$ and $\sigma_{n}$. The value of $(i,j)$ position in the $n_{th}$ Gaussian heatmap $G_{n}$ is shown in formula (\ref{eq:gaussian}).

\begin{equation}
  G_n(i,j) = e^{-\frac{(i-kr_n)^2+(j-kc_n)^2}{2\sigma_n^2}}
  \label{eq:gaussian}
\end{equation}

Then the proposed $RCC$ layer has following properties.

\begin{theorem}
If N=1, output of $RCC$ layer is $C_1$ times the input heatmap, where $C_1=\sum_{i=1}^W e^{-\frac{(i-kc_1)^2 + (i-kr_1)^2}{2\sigma_1^2}}$, W is width of the heatmap.
\label{th:theorem1}
\end{theorem}

\begin{theorem}
If $N>1$, output of RCC layer is  
\begin{equation}
  RCC(l,m) = \sum_{a \ne b}C_{a,b} G_{a,b}(l,m) + \sum_{n=1}^N C_{n}G_n(l,m)
\end{equation}
where $a,b\in \{1,2,...,N\}$ and 

\begin{align}
  &G_{a,b}(l,m)= e^{-\frac{\sigma_b^2(l-kr_a)^2 +\sigma_a^2(m-kc_b)^2}{2\sigma_a^2\sigma_b^2}} \\ 
  &C_{a,b}= \sum_{i=1}^W e^{-\frac{\sigma_b^2(i-kc_a)^2 + \sigma_a^2(i-kr_b)^2}{2\sigma_a^2\sigma_b^2}} \\
  &C_{n}= \sum_{i=1}^W e^{-\frac{(i-kc_n)^2 + (i-kr_n)^2}{2\sigma_n^2}}
  \label{eq:jianhua} 
\end{align}

\label{th:theorem2}
\end{theorem}

Theorem \ref{th:theorem1} indicates that the output of RCC can be same with the input Gaussian heatmap in single-peak situation by normalizing the output to [0,1]. Thus, the RCC layer does not affect the results of single-peak predictions. When there are multi-peaks in the heatmap, Theorem \ref{th:theorem2} shows that all the Gaussian including their joint Gaussian sub-heatmaps $G_{a,b}(l,m)$ are weighted, thereby enlarging the difference between these peaks. Generally speaking, those Gaussian sub-heatmaps with a small $\sigma$ and a $(kr_b, kc_a)$ or $(kr_n, kc_n)$ far from the center of the heatmap are assigned a low weight, as illustrated in (a), (b) of Fig. \ref{fig:spcases}. There are cases that the weights of some $G_n$ are equal. However, $C_{a,b}$ is often assigned a higher weight in this case, and the peaks of these joint sub-heatmaps tend to be on the diagonal. As shown in (c) of Fig. \ref{fig:spcases}, these peaks may be far from the ground truth and result in a big loss during the training. Thus, networks would be aware of the multi-peaks situation, and avoid it as the loss decreases. 

The output of RCC layer is normalized to [0,1] by a normalization factor on the basis of above analysis. This transformation makes the output of RCC remain the same as the input when $N=1$ and does not change the conclusion when $N>1$. The normalization is shown in (\ref{eq:norrcc}).

\begin{equation}
  \overline{RCC(l,m)} = \frac{\sum_i HT(l,i)HT(i,m)}{Z}
  \label{eq:norrcc}
\end{equation}
where Z is a normalization factor with a value that equals to the maximum in the output of RCC layer. 

Note that RCC is not only used during the inference, it needs to be introduced into training. Although the prior of this layer is not always right, its main purpose is to let the network know that single-peak output is better during training. Thus, the network is optimized towards more accurate and single-peak. In this way, the network can learn to avoid multi-peaks predictions.

\subsection{Highly differentiated heatmap regression}

The heatmap used as the ground truth is generally generated by a Gaussian distribution. The maximum of the heatmap is too close to the values of neighboring pixels, thus a slight change in the output may change the position of the predicted keypoint. In view of this problem, we propose a highly differentiated heatmap and a weighted cross-entropy loss.

\begin{figure}[tb]
   \centering
   \subfigure[TR. heatmap]{
      \includegraphics[width=0.40\linewidth]{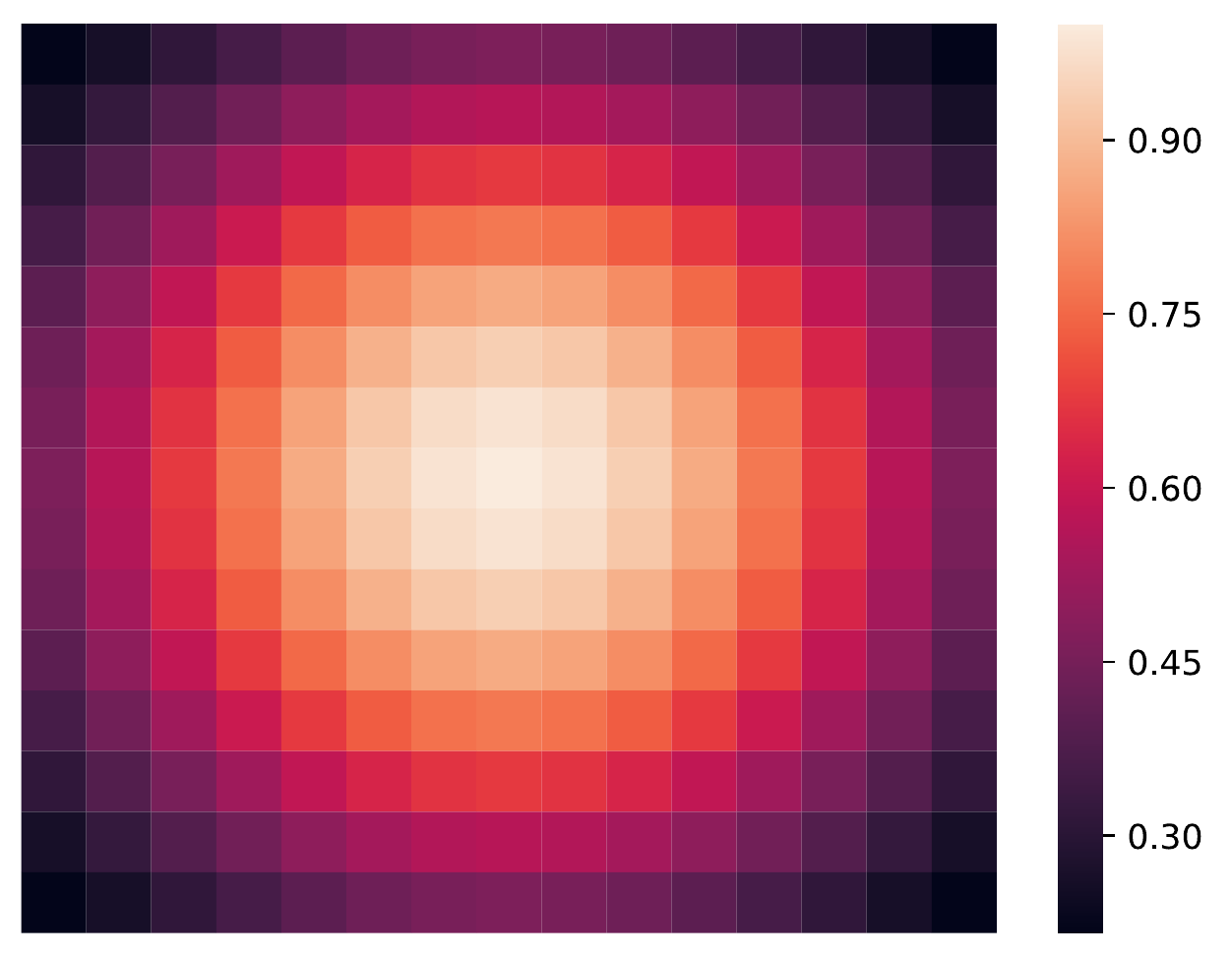}
   }
   \subfigure[HD. heatmap]{
      \includegraphics[width=0.40\linewidth]{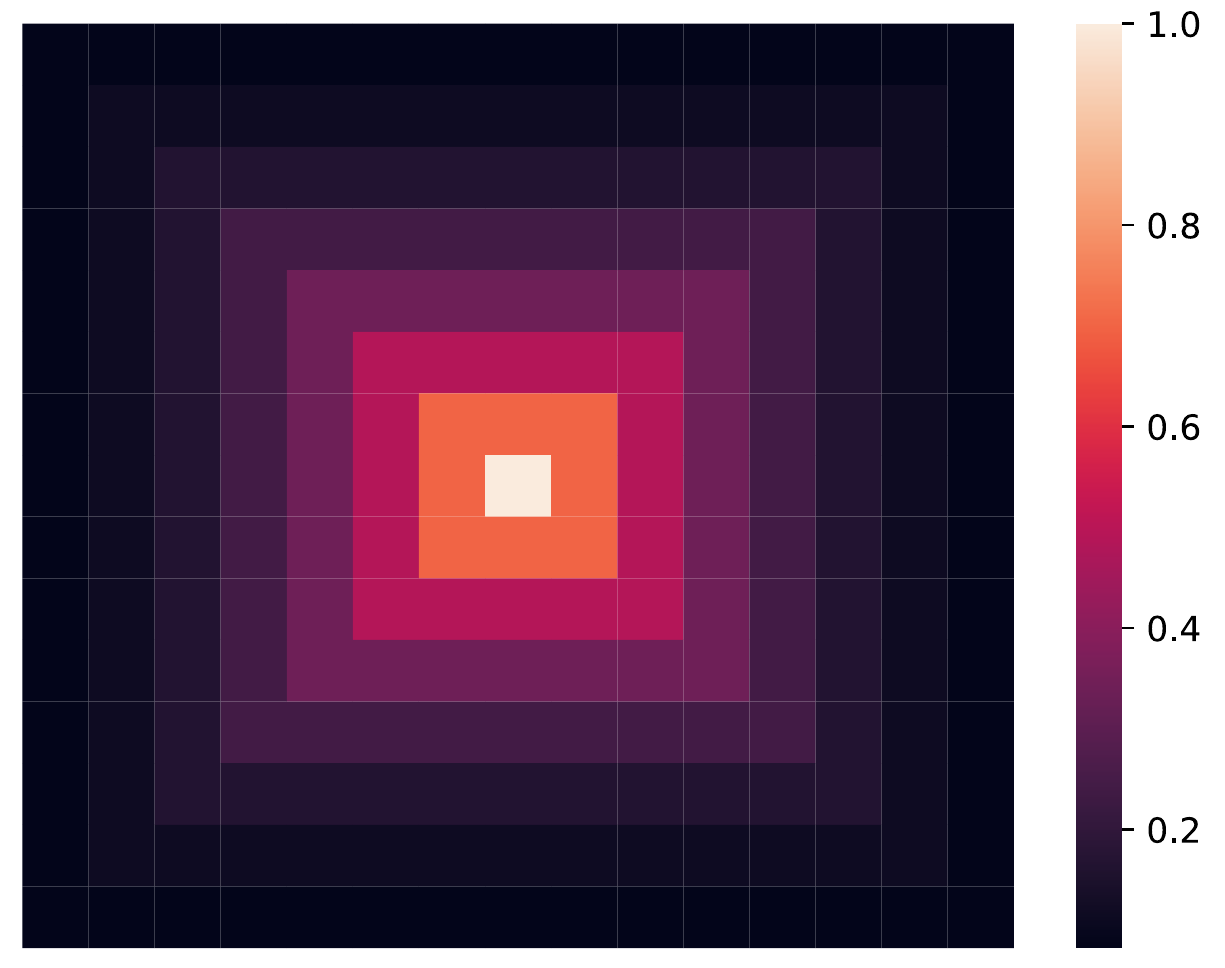}
   }
   \caption{The difference between the traditional heatmap (TR. heatmap) and highly differentiated heatmap (HD. heatmap). For generating HD. heatmap, we use the chessboard distance metric in this paper, but other distance metrics such as Euclidean or City Block distance could also be suitable choices. The hyperparameter $T_{hd}$ is set to 8 and $\alpha$ is set to 0.7 in the experiments.}
   \label{fig:hdheatmap}
\end{figure} 

\subsubsection {Highly differentiated heatmap} 

The highly differentiated heatmap is constructed as follows: The value of a keypoint's position is 1 in the heatmap, and the value of the point with a distance $d_p$ from the keypoint is $\alpha^{d_p}$, where $\alpha$ is the hyperparameter with a value in range (0, 1). When distance $d_p$ is greater than a certain threshold $T_{hd}$, the value of this point is set to 0. The highly differentiated heatmap increases the difference between keypoints and the other points in the heatmap, making the features of the keypoint more discriminative. The traditional heatmap and highly differentiated heatmap are shown in Fig. \ref{fig:hdheatmap}. 

\subsubsection {Weighted cross-entropy loss} It is difficult to converge when directly regressing the highly differentiated heatmap using L1 or L2 loss function. Since the features of the keypoint are similar to that of the surrounding points, the intention to enlarge the difference between the keypoint and its surrounding points leads to confusion of the network. We solve this problem by regarding the heatmap regression task as a multi-label classification. A multi-label map is built as the annotations. The number of pixels in the highly differentiated heatmap is regarded as the number of categories in the multi-label map. The points in the highly differentiated heatmap with value 0 mean the input does not belong to this category and the others represent the input has these category labels. The schematic diagram of multi-label map can be found in Fig. \ref{fig:MPT}. The multi-label map provides an area where a keypoint should be in. Commonly used loss function for multi-label classification is the cross-entropy, but the use of cross-entropy with the multi-label map cannot increase the discrimination between the keypoint and its surrounding points. Thus, we propose a weighted cross-entropy loss, defined in the equation (\ref{eq:weight_loss}), to solve this problem.

\begin{equation}
   \mathcal L_{wce} = - \sum_{x \in I} \sum_{i,j} w_x^{i,j} y_x^{i,j} log r_x^{i,j}
   \label{eq:weight_loss}
\end{equation}
where $x$ is an image from the training set $I$, $(i, j)$ represents the position of column $j$ of the $i_{th}$ row. $w_x^{i,j}$ is the value of $(i, j)$ position in highly differentiated heatmap and $y_x^{i,j}$ is the label of the $(i, j)$ position in multi-label map. $r_x^{i,j}$ is the value of $(i, j)$ position in the output heatmap. The loss means that the pixel value of the highly differentiated heatmap is used as the weight of the cross-entropy loss. At the same time, we use a softmax function on the output heatmap before calculating the loss. The softmax operation increases the competitive relationship between different pixels, where the position of higher weight, the greater the predicted value. Thus, the distance between maximum and the second-largest value in the heatmap is enlarged.

\subsection{Maximum stability training}

In order to make the network more robust to unseen natural perturbations, we propose a maximum stability training loss to explore the relation between a clean image and its perturbed image. A typical approach is to make the outputs of them similar, also known as stability training \cite{zheng2016improving}. However, this loss is too strict for pose estimation. The output heatmap represents the probability that each pixel in the image belongs to a keypoint, and each pixel of a clean image may be changed in its perturbed image. Forcing the probability of all the pixels between these image pairs to be the same increases difficulty in network optimization. 

In fact, the change in the maximum of a heatmap contributes most to stability. As the keypoint is the position with max probabilities in the heatmap, if the maximum of an image and that of its perturbed image remain in the same position, the predicted keypoint would be stable. Based on above analysis, we propose a new regularization loss, with the change of maximum being focused on. 

\begin{equation}
   \mathcal L_{mst} = \sum_{x \in I, x' \in I_{x}^{mp}} {\frac{\sum_{i,j}{(\bigtriangleup r_{x}^{i,j})^2}}{H \cdot W}  + (\bigtriangleup r_{x}^{(1)})^2}
   \label{eq:regularizationloss}
\end{equation}
where $r_x^{i,j}$ is the value at $(i,j)$ position of an output heatmap $r_x$ with an input x. $\bigtriangleup r_x^{i,j} = r_{x'}^{i,j}-r_x^{i,j}$, $x'$ is a perturbed image of $x$ and is obtained from pre-defined multiple perturbations $I_x^{mp}$. $H,W$ are the height and width of a heatmap respectively. $\bigtriangleup r_{x}^{(1)}$ is the difference between maximum of $r_x$ and the value of same position in $r_{x'}$. This loss strengthens the limit on the change of $r_x^{(1)}$, allowing small changes in other positions, thereby reducing the difficulty of optimization. Meanwhile, we provide a theoretical certification to demonstrate that our maximum stability training loss can indeed improve robustness. 

\begin{theorem}
  In the same setting in (\ref{eq:regularizationloss}), with $r_x^{(1)}$ and $r_x^{(2)}$ being the first and second largest values in the output heatmap $r_x$.

If the following condition is satisfied,
\begin{center}
$r_x^{(1)} - r_x^{(2)} \ge \sqrt{\sum_{i,j}{(\bigtriangleup r_{x}^{i,j})^2} + H\cdot W(\bigtriangleup r_x^{(1)})^2}$
\end{center}

then,
\begin{center}
$argmax_{i,j}(r_x^{i,j}) = argmax_{i,j}(r_{x'}^{i,j})$.
\end{center}
\label{th:theorem3}
\end{theorem}

This property manifests that a larger $r_x^{(1)} - r_x^{(2)}$ and a smaller $\sqrt{\sum_{i,j}{(\bigtriangleup r_{x}^{i,j})^2} + H\cdot W(\bigtriangleup r_x^{(1)})^2}$ are beneficial for improving the stability. $r_x^{(1)} - r_x^{(2)}$ is already enlarged by HDHR. Therefore, minimizing the proposed loss ${\frac{\sum_{i,j}{(\bigtriangleup r_{x}^{i,j})^2}}{H\cdot W} + (\bigtriangleup r_x^{(1)})^2}$ could further enhance the robustness.

\section{Experiments}

\subsection{Datasets}

We conduct experiments on two benchmark datasets of hand pose estimation-RHD \cite{zimmermann2017learning} and STB \cite{zhang2017hand}. RHD is a synthetic hand pose dataset. It contains 41,258 training images and 2,728 test images. Precise 2D annotations for 21 keypoints are provided. STB is a real-world dataset. It contains 12 sequences with six different backgrounds. Following same setting in \cite{yang2019aligning,zimmermann2017learning}, 10 sequences are used for training and the other 2 for testing.

There are lots of differences between synthetic datasets and real datasets in terms of background, annotation mechanism and so on. Thus, using these two kinds of datasets allows for comprehensive evaluation of algorithms. 

\subsection {Experimental setup}
\label{sec:heatmapsetting}
Following the creation of perturbed images in \cite{hendrycks2019benchmarking}, we generated 15 kinds of perturbations-brightness, defocus blur, zoom blur, frost, contrast, gaussian noise, glass blur, motion blur, shot noise, snow, gaussian blur, jpeg compression, saturate, spatter and speckle noise-each with five levels of severity. Since the perturbations in practical applications cannot all be seen during training, we train the network with first 10 kinds of perturbations and evaluate on the others. Each network used in our evaluation consists of two parts-a backbone network and an upsampling network. The backbone network is used to extract latent features of an input image. The upsampling network utilizes the features to predict the heatmaps.

\subsection{Evaluation criteria}

\textbf{Accuracy} The area under the curve (AUC) on the percentage of correct keypoints are used to measure the performance of pose estimation.

\textbf{Robustness} Different from the classification task, the robustness of a regression task has little connect to accuracy \cite{zhang2016stability}. Thus, we define a new criteria $R(T, T^{mp})$ to compare the robustness of different models, where $T^{mp}$ is a perturbation set of validation set $T$. The definition of $R(T, T^{mp})$ is shown in (\ref{eq:robindex}).

\begin{equation}
  R(T, T^{mp}) = \sum_{x \in T}\sum_{x' \in T_{x}^{mp}}(Pos(r_x^{(1)}) - Pos(r_{x'}^{(1)}))^2
  \label{eq:robindex}
\end{equation}
where $T_{x}^{mp}$ is a set that contains all the perturbed images of $x$. $Pos(r_x^{(1)})$ and $Pos(r_{x'}^{(1)})$ represent the predicted keypoints of $x$ and $x'$ respectively.

After calculating the $R(\cdot, \cdot)$, the proportion of stable samples, where the predictions of the perturbed images in $T^{mp}$ remain the same with their natural images, are calculated. The predicted keypoints of an original image and that of its perturbed image are considered to be the same when the difference between them does not exceed $P \in N^+$ pixels ($N^+$ is a set of natural numbers). By taking different $P$, a Robustness Under Curve (RUC) can be obtained in a similar way to Accuracy Under Curve.

\subsection{Ablation study}

The proposed method can be divided into three parts, Row-Column Correlation (RCC), Highly Differentiated Heatmap Regression (HDHR) and Maximum Stability Training (MST). MobilenetV2 \cite{sandler2018mobilenetv2} is adopted as the backbone to study the role of the three modules. We first conduct experiments to demonstrate that RCC could alleviate the multi-peaks problem and HDHR can enlarge $r_x^{(1)} - r_x^{(2)}$. For evaluating RCC, we use a metric $D_n$, which represents the difference of positions between $top_n$ and $top_1$ in the validation set $T$, as shown in (\ref{eq:dn}).
\begin{equation}
  D_n(T) = mean(\sum_{x \in T}\sum_{i=2}^n (Pos(r_x^{(i)}) - Pos(r_x^{(1)}))^2)
  \label{eq:dn}
\end{equation}
where $r_x^{(i)}$ is the $i_{th}$ largest value in the heatmap and $Pos(r_x^{(i)})$ represents its position. The smaller the $D_n$, the more concentrated the high response of a heatmap. As for HDHR, a metric (\ref{eq:d12}) is adopted. 
\begin{equation}
D_{(1)(2)}(T) =mean(\sum_{x \in T} r_x^{(1)} - r_x^{(2)})
\label{eq:d12}
\end{equation}

\begin{figure}[htbp]
  \centering
  \includegraphics[width=0.97\linewidth]{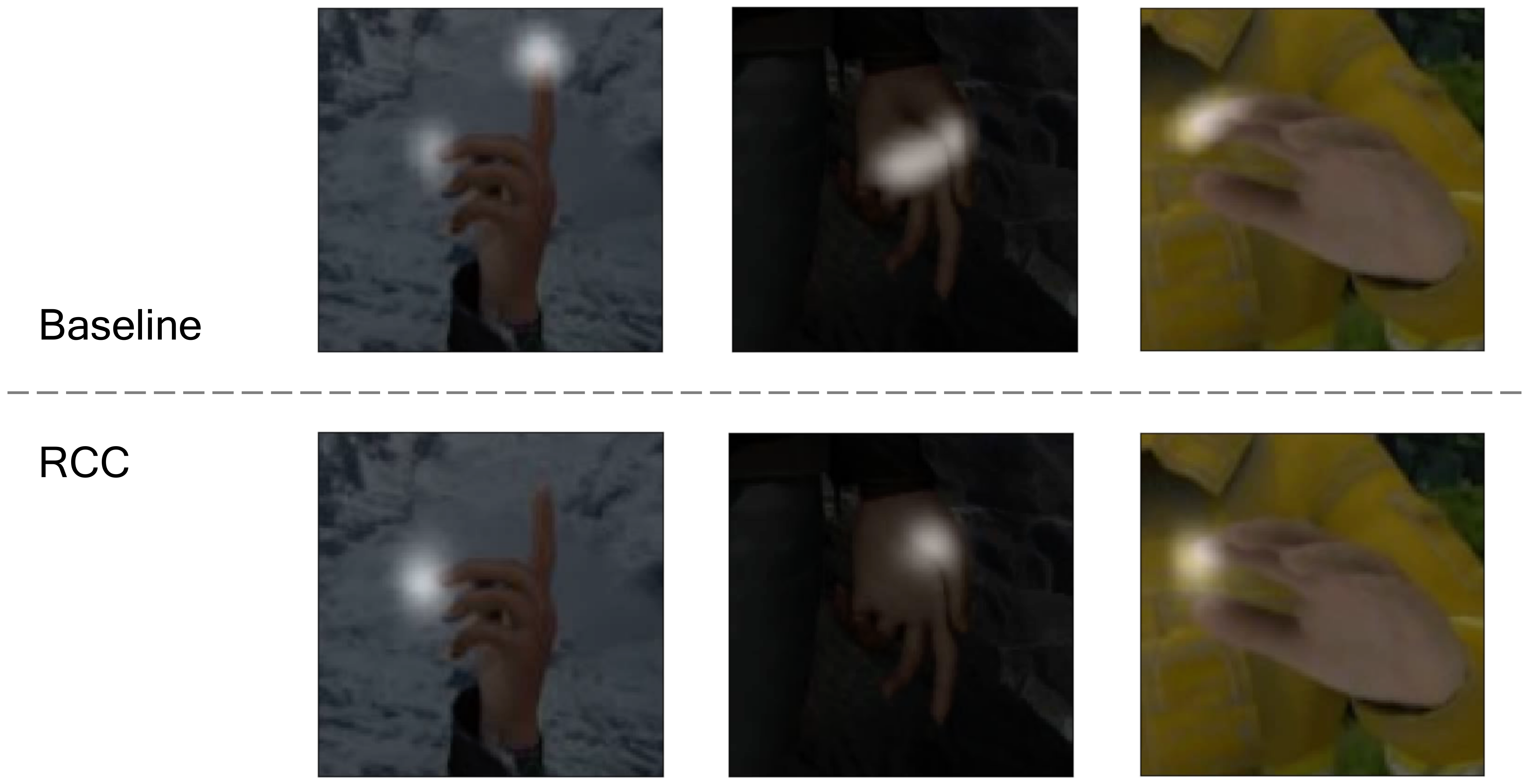}
  \caption{The visual comparisons between the predicted heatmaps of the Baseline model and the RCC model. }
  \label{fig:rcc_output}
\end{figure}

The results are shown in Table \ref{tab:dn12}. Baseline represents the traditional heatmap regression. RCC achieves a smaller $D_n(T)$ and HDHR enlarges the $D_{(1)(2)}(T)$ in both RHD and STB datasets. The effectiveness of both modules has been verified. In addition, we show the predicted heatmaps of the RCC model and the baseline model in Fig. \ref{fig:rcc_output}. Baseline model tend to output heatmaps with multi peaks while the RCC model tend to output single peak heatmaps. This phenomenon demonstrates that RCC could alleviate multi-peaks problem in real-world evaluation.

\begin{table}[tbp] 
  \setlength{\tabcolsep}{1.7mm}{
   \begin{tabular}{ccc|ccc} 
    \toprule 
    $D_n (\downarrow)$ &  RHD & STB & $D_{(1)(2)} (\uparrow)$ & RHD & STB \\
    \hline 
    Baseline  & 954.67 & 820.22 &  Baseline & 0.048 & 0.055 \\  
    RCC & \textbf{870.67} & \textbf{749.68} & HDHR & \textbf{0.099} & \textbf{0.203} \\ 
    \bottomrule 
   \end{tabular}
   \caption{Experiments for verifying the effectiveness of RCC and HDHR. The n is set to 64 when calculating $D_n$. 
   } 
   \label{tab:dn12} 
   }

  \end{table}

\begin{figure}[tbp]
  \centering
  \includegraphics[width=0.99\linewidth]{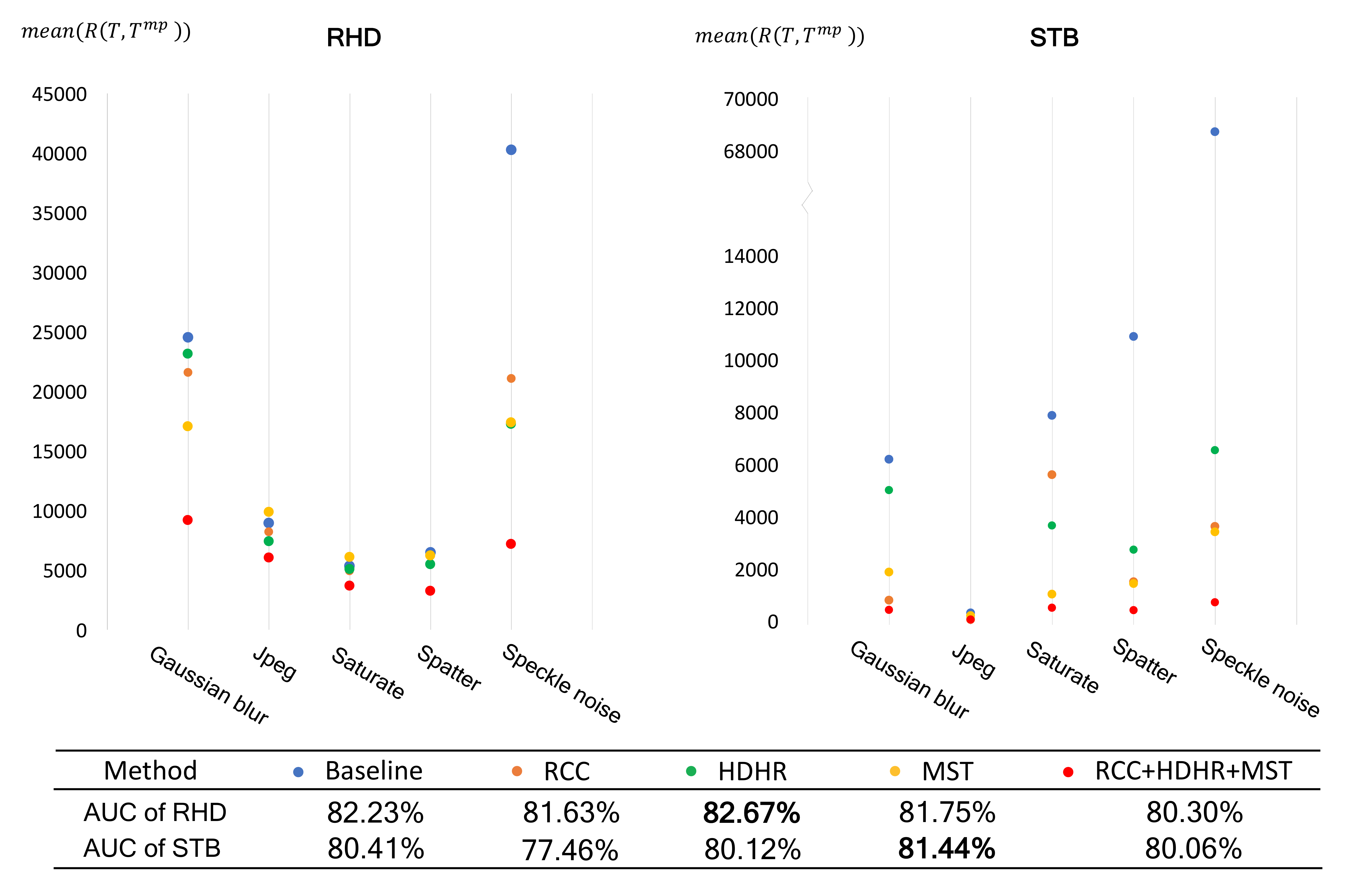}
  \caption{Robustness of different training strategies on RHD and STB datasets. The ordinate is the average value of $R(T, T^{mp})$ ($\downarrow$ is better) of each perturbation and abscissa shows the name of perturbations. AUC ($\uparrow$ is better) is calculated with thresholds 0-30 pixel on clean images.}
  \label{fig:mob_rhp_perturb}
\end{figure}

\begin{figure}[htbp]
  \centering
  \subfigure[RHD]{
     \includegraphics[width=0.47\linewidth]{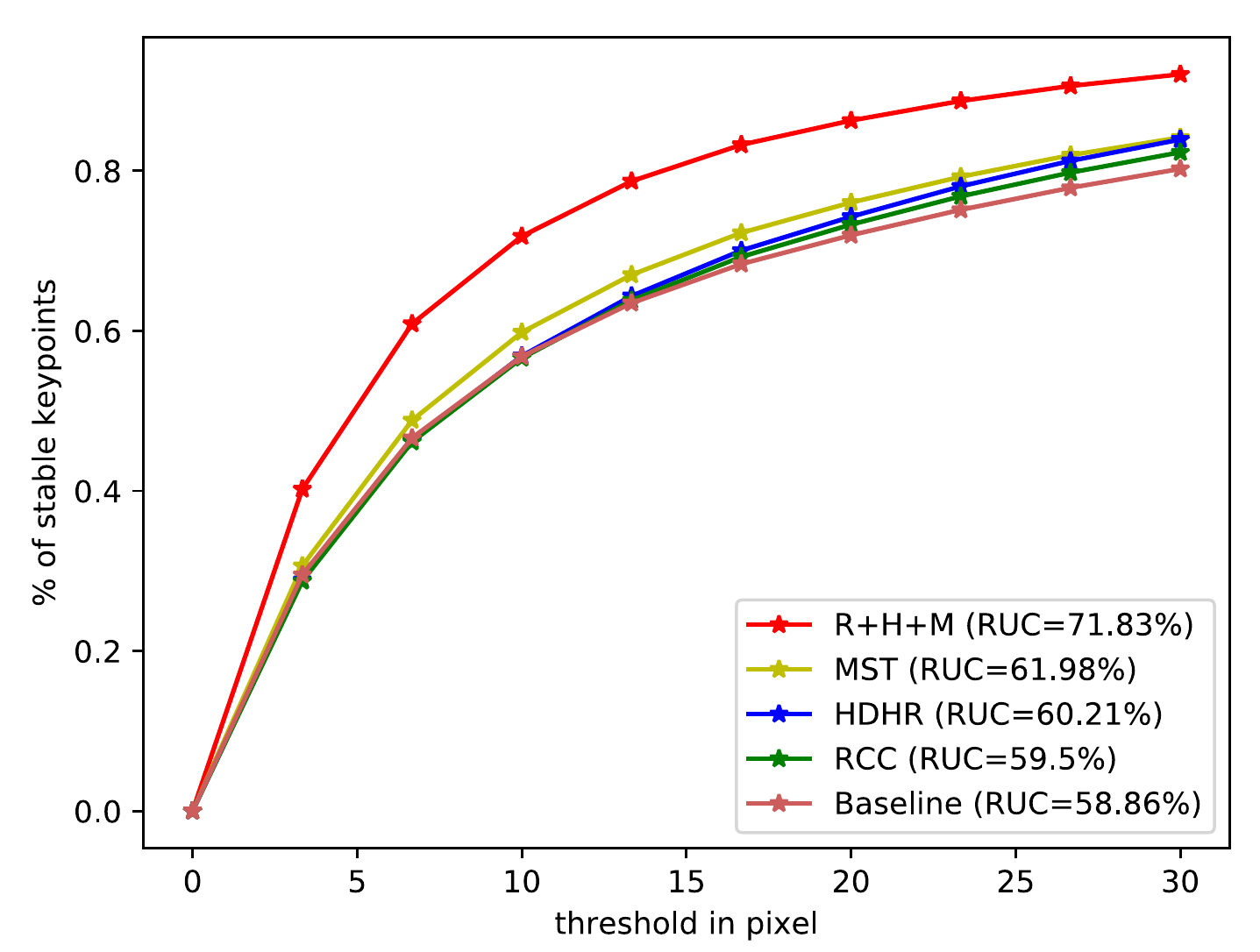}
     \label{fig:ruc_ablation_rhd}
  }
  \subfigure[STB]{
     \includegraphics[width=0.47\linewidth]{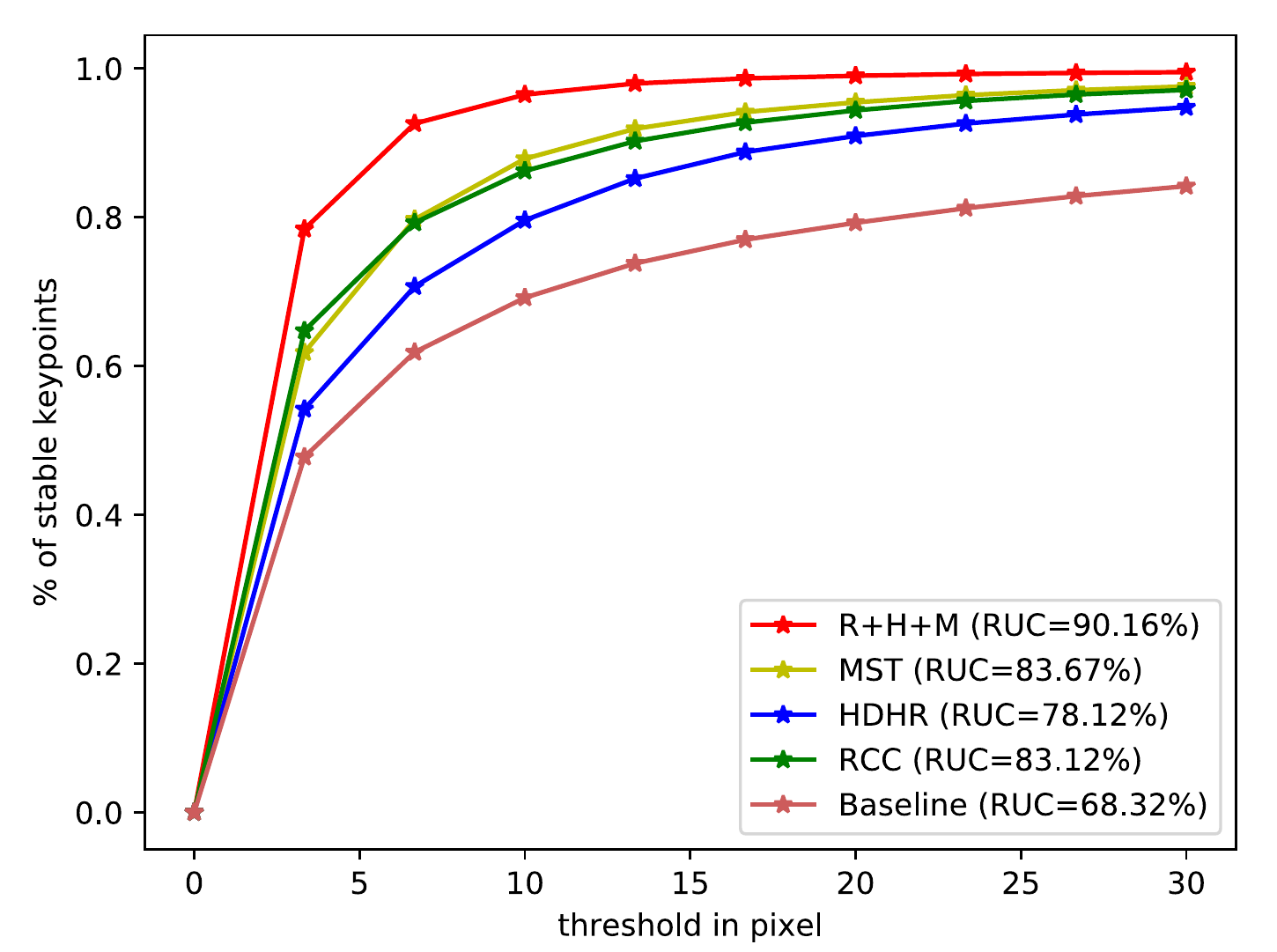}
     \label{fig:ruc_ablation_stb}
  }
  \label{fig:all_experiemnts1}
  \caption{RUC ($\uparrow$ is better) of different training strategies. R+H+M is an abbreviation of RCC+HDHR+MST.}
\end{figure}

$R(T, T^{mp})$ and AUC of the three modules is shown in Fig. \ref{fig:mob_rhp_perturb} and RUC in Fig. \ref{fig:ruc_ablation_rhd} and \ref{fig:ruc_ablation_stb}. Although there are many difference between RHD and STB, where RHD is a synthetic dataset and STB is a real-world dataset, robustness against five perturbations are enhanced after using RCC, HDHR and MST. However, $R(T, T^{mp})$ of MST on RHD is slightly worse than Baseline on Jpeg and Saturate perturbations. As shown in Theorem \ref{th:theorem3}, the robustness improvement of MST loss is related to $r_x^{(1)} - r_x^{(2)}$. However, traditional heatmap regression has tiny $r_x^{(1)} - r_x^{(2)}$ because Gaussian heatmap is used as the ground truth. And the $r_x^{(1)} - r_x^{(2)}$ of RHD is smaller than that of STB (Table \ref{tab:dn12}), thus only using $\mathcal L_{mst}$ is unable to reach its full potential
on RHD dataset. When MST is combined with HDHR, the robustness is significantly enhanced. In addition, the drop of AUC on clean images is reasonable, since many papers in the field of adversarial examples have demonstrated that the improvement in the robustness is accompanied by the decrease in the accuracy of the clean images \cite{zhang2019theoretically}.

\subsubsection{Different model architectures}

\begin{figure}[tbp]
  \subfigure[RHD]{
    \includegraphics[width=0.47\linewidth]{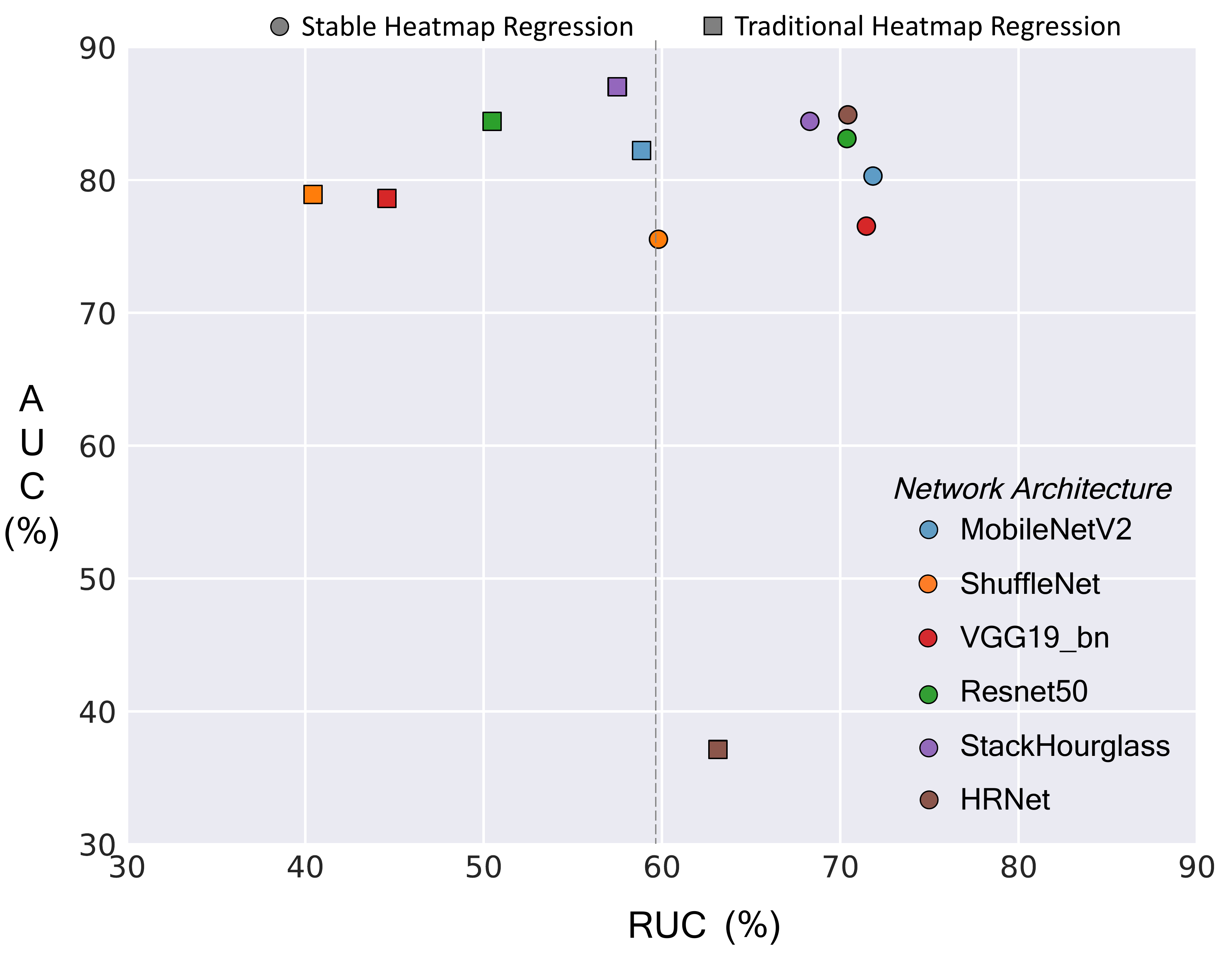}
    \label{fig:rhdrob}
 }
 \subfigure[STB]{
  \includegraphics[width=0.47\linewidth]{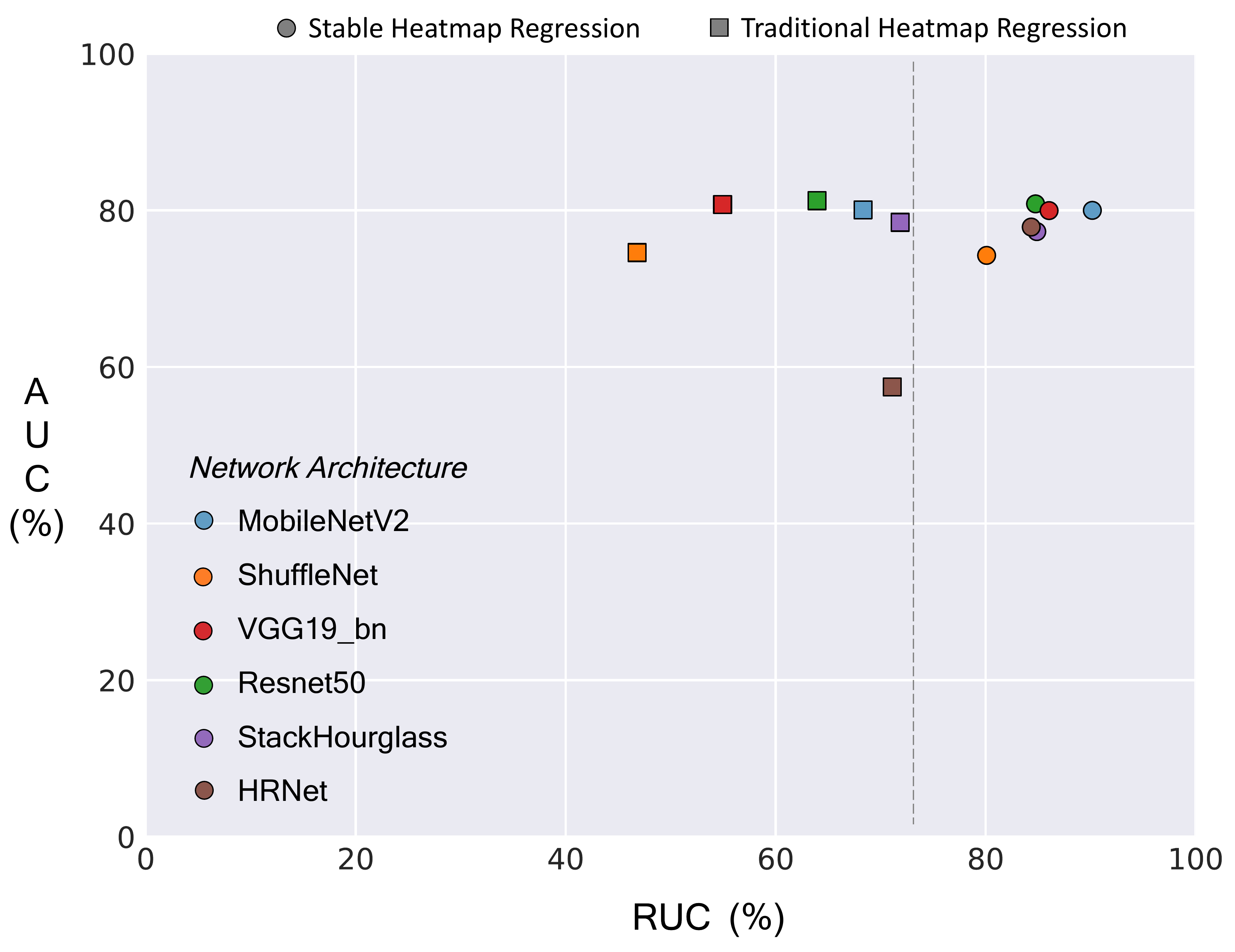}
  \label{fig:stbrob}
}
\caption{AUC ($\uparrow$ is better) and RUC ($\uparrow$ is better) of different model architectures.}
\label{fig:all_experiemnts2}
\end{figure}

Different backbones of pytorch official realization, resnet50 \cite{he2016deep}, vgg19\_bn \cite{simonyan2014very}, mobilenetV2 \cite{sandler2018mobilenetv2}, shufflenet \cite{ma2018shufflenet}, stackhourglass (2-stack) \cite{newell2016stacked} and HRNet \cite{wang2020deep} are adopted for comprehensive evaluation. The results on RHD dataset are shown in Fig. \ref{fig:rhdrob} and STB in Fig. \ref{fig:stbrob}. Our method achieves most robust models against small perturbations and maintains high performance in both datasets on all models. The experimental results demonstrate the potential that the proposed method can be a general substitute for traditional heatmap regression methods.

\subsection{Comparison with state of the art}

\begin{figure}[h]
  \subfigure[RHD]{
    \includegraphics[width=0.47\linewidth]{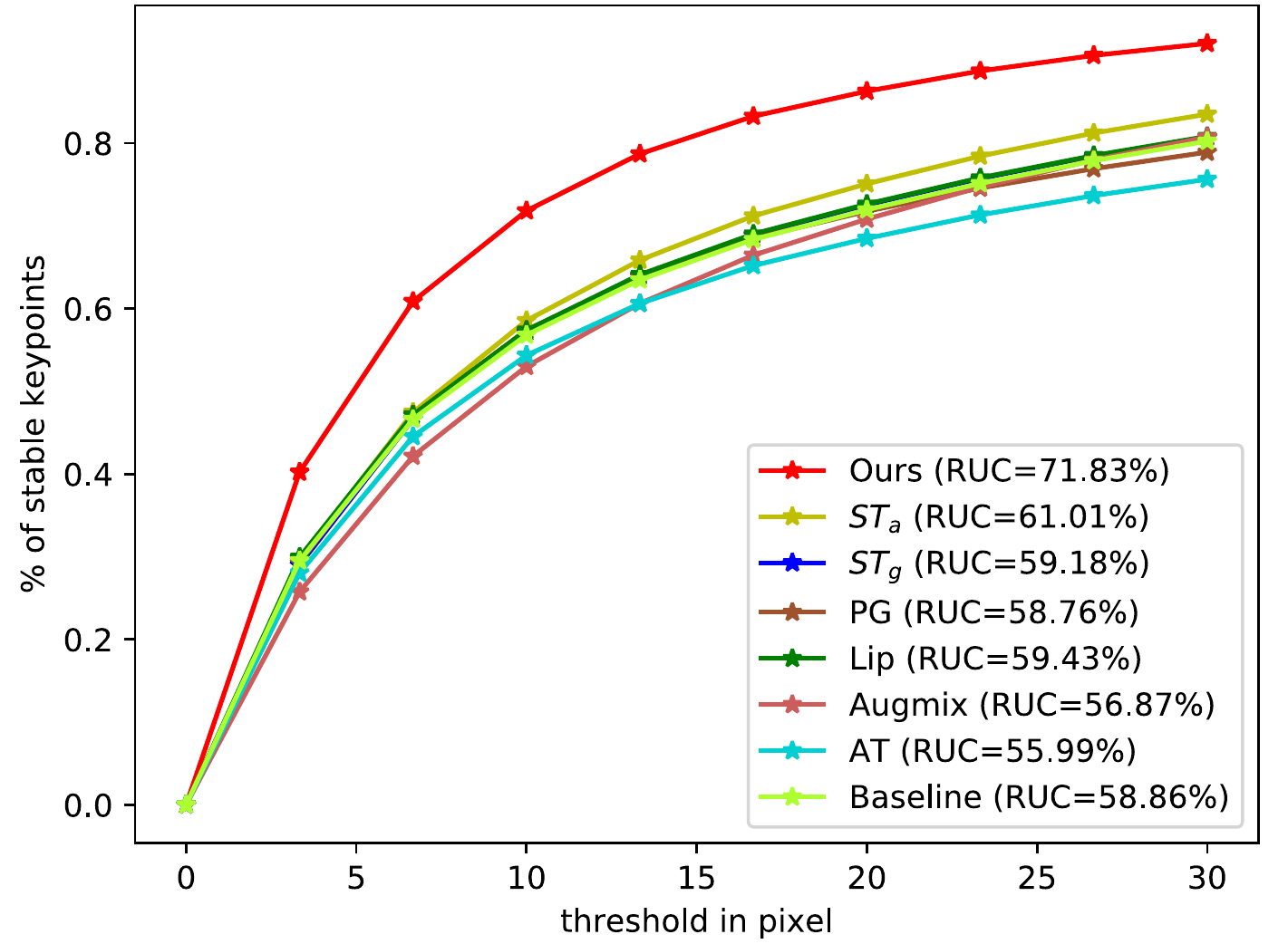}
    \label{fig:state_art_rhd}
  }
\subfigure[STB]{
  \includegraphics[width=0.47\linewidth]{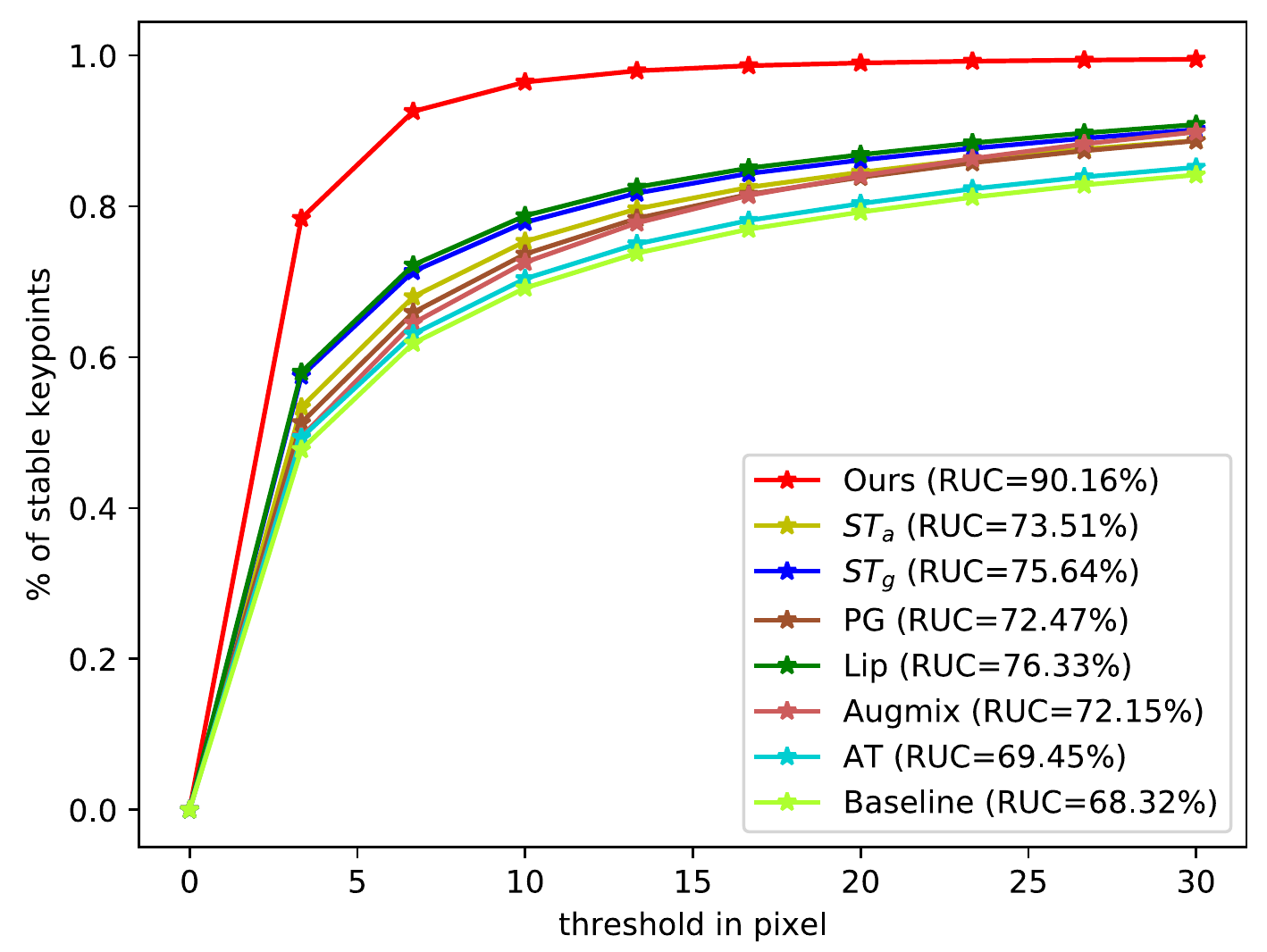}
  \label{fig:state_art_stb}
}
  \caption{RUC comparisons with SOTA methods. The backbone is mobilenetV2.}
\label{fig:all_experiemnts3}
\end{figure}

The proposed method is compared with state-of-the-art (SOTA) methods for improving robustness, AT \cite{madry2017towards}, AugMix \cite{hendrycks2019augmix}, Lip \cite{usama2018towards}, PG \cite{lopes2019improving} and ST \cite{li2019certified,zheng2016improving}. Note that ST \cite{zheng2016improving} added a stability loss $\mathcal L_{stability} = \parallel r_x - r_{x'} \parallel$ to make the outputs of clean images and their perturbed images stable and \cite{li2019certified} proved that using gaussian noise to construct perturbed images is sufficient to improve robustness against multiple perturbations. However, several papers \cite{tramer2019adversarial,yin2019fourier} demonstrated that using a diverse set of augmentations is necessary to improve the robustness. Therefore, we use all the perturbations in the training of MST, including the gaussian noise, to calculate the stability loss $\mathcal L_{stability}$ as a supplementary method. This modified version of ST is denoted as $ST_a$ and the original version is $ST_g$. RUC results of these methods on RHD and STB are shown in Fig. \ref{fig:state_art_rhd} and Fig. \ref{fig:state_art_stb}, and AUC in Table \ref{tab:state_art}. The proposed method achieves an advance over these SOTA methods as well as maintains high performance on RHD and STB datasets. In addition, RUC of MST in Fig. \ref{fig:ruc_ablation_rhd} and Fig. \ref{fig:ruc_ablation_stb} is better than that of $ST_a$ and $ST_g$ in Fig. \ref{fig:state_art_rhd} and Fig. \ref{fig:state_art_stb}, which manifests that focusing on the change of maximum is more efficient in pose estimation. 

\begin{table}[htbp]
  \centering
  \setlength{\tabcolsep}{0.5mm}{
  \begin{tabular}{cccccccccc}
    \toprule
    AUC(\%) &  Base. & AT & Aug. & Lip & PG & $ST_g$ & $ST_a$ & \textbf{HDHR} & \textbf{Ours} \\
    \midrule
    RHD    & 82.2  & 81.2 & 82.1 & 79.7 & 80.3 & 79.5 & 80.4  & 82.6 & 80.3 \\
    STB    & 80.4  & 79.3 &  80.9 & 81.5 & 80.1 & 79.6 & 80.0 & 80.1 & 80.0 \\ 
    \bottomrule
  \end{tabular} 
  }
  \caption{AUC of state-of-the-art methods and the proposed method. Base. and Aug. are the abbreviations of Baseline and Augmix. The backbone is mobilenetV2. } 
  \label{tab:state_art}  
  \end{table}

\section{Discussion}
\subsection{AUC under different image corruptions}

The test samples provided by public datasets are relatively clean samples, and the test samples in the real-world applications have many image corruption problems. Therefore, a higher AUC achieved with the test samples of the public datasets does not mean a better visual effect in real-world applications. We evaluate the AUC results of the proposed and the corresponding baseline models under different corruption samples, and find that models with a high RUC have better accuracy results on the corruption samples. The results are shown in Table \ref{tab:AUC_dif}. In fact, our method has better visual performance even with the original test samples since the keypoints predicted by our method are more stable than those predicted by the baseline model. 

\begin{table}[h]
  \centering
  \setlength{\tabcolsep}{0.5mm}{
  \begin{tabular}{cccccccc}
    \toprule
    \multicolumn{2}{c}{AUC(\%)} &  Original & GB & Jpeg & Saturate & Spatter & SN\\
    \midrule
    \multirow{2}*{RHD} & Baseline    & 82.23 & 51.48 & 66.29 & 73.37 & 69.69 & 41.95  \\ 
    & Ours    &  80.30  &  64.80 & 67.82 & 72.94 & 73.41 & 66.99 \\ 
    \midrule
    \multirow{2}*{STB} & Baseline  & 80.41  & 59.58 & 77.68 & 53.40  & 54.91 & 29.03 \\ 
    & Ours    & 80.06  &  59.79 & 77.16 & 59.82  & 61.73 & 45.21 \\
    \bottomrule
  \end{tabular} 
  }
  \caption{AUC of the proposed models and the baseline models under different corruptions on RHD and STB dataset. The backbone is mobilenetV2. "GB" is the abbreviation of gaussian blur and "SN" the speckle noise.}
  \label{tab:AUC_dif}  
  \end{table}

\subsection{Loss surface}

\begin{figure}[htp]
  \centering
  \includegraphics[width=0.97\linewidth]{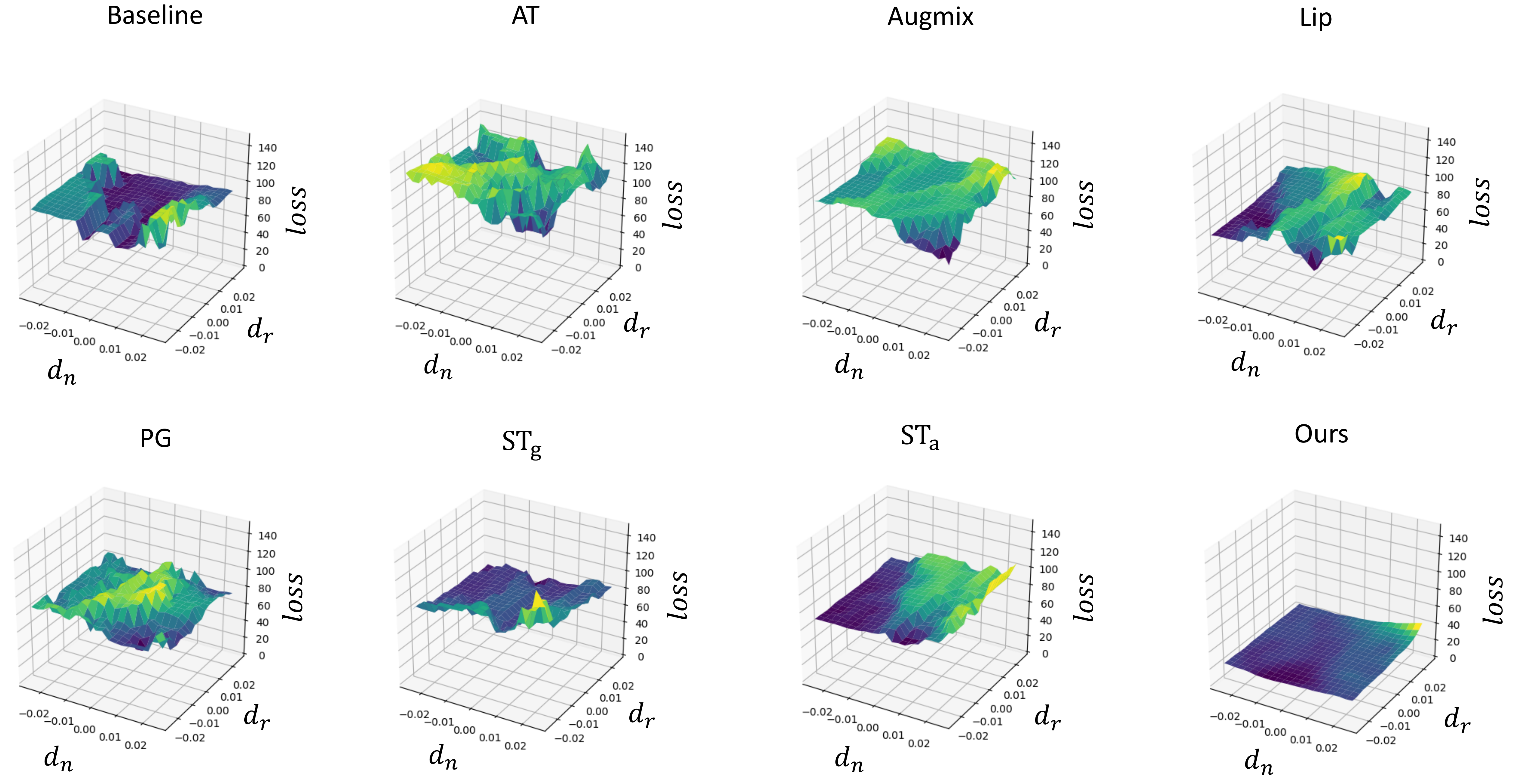}
  \caption{Loss surfaces of our method and SOTA methods on RHD dataset.}
  \label{fig:loss_surface}
\end{figure}

We plot loss surfaces around test data points in Fig. \ref{fig:loss_surface}. We vary the input along a linear space defined by a direction of a difference vector ($d_n$) between a perturbed image and its clean image and a direction of a Rademacher vector ($d_r$), where the x and y-axes represent the magnitude of the perturbation added in each direction and the z-axis represents the loss. Our method achieves smoothest loss surface, demonstrating that a true boost for robustness is gained. 

\section{Conclusions}

In this paper, we propose a stable heatmap regression method to improve robustness for pose estimation models. The method alleviates the multi-peaks problem, makes the keypoint discriminative in the heatmap and applies a suitable stability loss for pose estimation, thereby achieving a robust pose estimation model without being vulnerable to small perturbations. The effectiveness of the method is validated by theoretical analysis and extensive experiments on two benchmark datasets with different model architectures. In the future, we will construct multi-perturbation sets more effectively to further improve robustness.

{\small
\bibliographystyle{ieee_fullname}
\bibliography{egbib}
}

\end{document}